# Marvin: Semantic annotation using multiple knowledge sources

Nikola MILOSEVIC[a]
[a] *School of Computer Science, University of Manchester*

**Abstract:** People are producing more written material then anytime in the history. The increase is so high that professionals from the various fields are no more able to cope with this amount of publications. Text mining tools can offer tools to help them and one of the tools that can aid information retrieval and information extraction is semantic text annotation. In this report we present Marvin, a text annotator written in Java, which can be used as a command line tool and as a Java library. Marvin is able to annotate text using multiple sources, including WordNet, MetaMap, DBPedia and thesauri represented as SKOS.

## 1. Introduction and background

In the past three decades, the amount of content available on the internet overcomes the amount of content published on paper in human history. Estimation is that around 50 million scholarly articles were published in our history (Jinha, 2010). Google estimated that there are more than 129 million published books in the World (Taycher, 2010), while there are about 4.6 billion web pages in the indexed web and nearly 550 billion individual documents in non-indexed web (Bergman, 2001). Projections say that the amount of data generated on the web will increase by 40% annually (Larose, 2014). World Wide Web provided people a novel ways to express themselves on a variety of media such as Wikis, question-answer databases, blogs, forums, review-sites and social media. It also provided accessible information and knowledge to people around the World. Almost everything can be found on World Wide Web. However, most of this information is in in textual form. Even corporation hold more than 80% of their information in textual manner (Grimes, 2008). Textual information is useful for people, since people are able to infer knowledge from the text, but machines are not able to do the same. In order to enable machines to process information, they need to be well structured. However, it is hard to structure textual information in old fashioned way into a classes or database tables. Information in text and in general information that describes the world needs more flexible approach. Semantic web and Linked Data emerged as a framework to enable describing a World. It provides flexibility by the use of triples. These triples contain semantics as a simple sentence containing subject, predicate and object. Subject and object are the things in the World, while predicate is describing the relationship between these things. Since linked data already have sentence-like structure, information from text can be transformed into linked data. Text can be seen as a way of describing a World understandable for humans, while linked data is the same process, but it is also understandable for machines. However, it is not easy to transform text into the linked data, because it requires a lot of text mining and natural language processing.

The main goal of linked data is to describe the world and to provide an easy framework for data integration. Probably the major source of knowledge currently published as a linked data is DBPedia, linked data version of Wikipedia. There are also other knowledge sources such as some specialised knowledge sources and ontologies that can be used on semantic web.

However, since the linked data is relatively new discipline, there are a vast amount of knowledge sources that could be used to add semantics, but are not transformed into linked data. Such resource is for example WordNet and for biomedical domain UMLS and MetaMap, which gives UMLS annotations to the given text.

Semantics of most of the texts cannot be annotated by a single semantic knowledge source. Especially this is true, if the text contains vocabulary from the particular domain. In this case, it is necessary to integrate knowledge from the general domain with the particular domain. This can be done by data integration of multiple knowledge sources.

In this work we present Marvin program that annotates text with annotations from various knowledge sources, both linked and non-linked data sources.

**2. Methods and Implementation**

Marvin program is implemented in Java and can be used both as a library or standalone application. It has its main function which enables it to be run as a standalone program, but also methods can be used as libraries, since annotation methods are made public. Currently, Marvin supports annotations using WordNet, MetaMap, DBpedia and custom SKOS thesauri. Marvin can annotate some text using all four knowledge sources, or any combination of them, which can be configured in configuration file (settings.cfg).

The general workflow of Marvin is shown on Figure 1. Text is firstly tokenized. Tokenization is performed using OpenNLP (Baldridge, 2005) and trained MaxEnt model provided by OpenNLP.

After the tokenization, annotation over the tokens is performed. However, for each knowledge source, the annotation is performed slightly different and for some knowledge sources additional normalizations. For example DBpedia annotation needs bigrams and trigrams to be generated and WordNet annotation needs word sense disambiguation. UMLS tagging also needs bigrams and trigrams and word sense disambiguation, but MetaMap already does this processing, so there was no need for implementing it.

To run Marvin semantic text annotator, it is advisable to have all necessary knowledge sources available on a local machine or local network, since that way it would not be subject of public server's blocks and a delays.

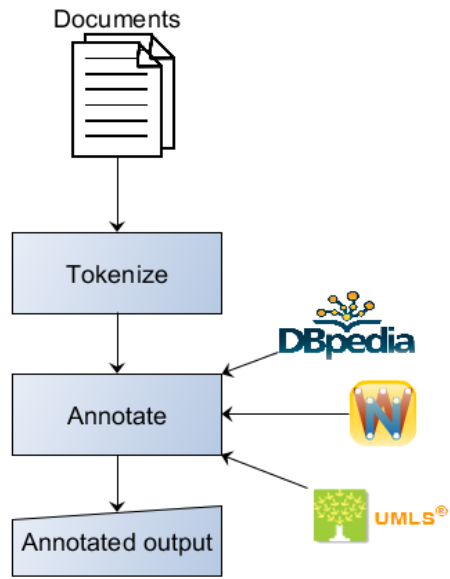

**Figure 1.** Overview of the workflow

*2.1. Annotating using DBpedia*

When annotating using DBpedia, our approach is to generate unigrams, bigram and trigrams from the supplied text. The rationale is that there are a number of definitions on Wikipedia and DBpedia that are one, two or three words long. After unigram, bigrams and trigrams are generated; we capitalize the first letter, since *labels* of DBpedia items are always with the first capital letter. Also, our approach puts the rest of the text in lowercase. After this, we query DBpedia with the generated strings. Querying is performed over the SPARQL interface. For testing we used public DBpedia interface (http://dbpedia.org/sparql). However, this interface has certain restriction on the number of queries that can be submitted. In case of larger texts, the interface may block the IP from which the requests are coming, which will result in 503 HTTP responses. For larger texts, it is advisable to have local instance of DBpedia and its SPQRQL interface.

The workflow diagram for annotation using DBpedia and its SPARQL interface is presented in Figure 2.

# Technical report

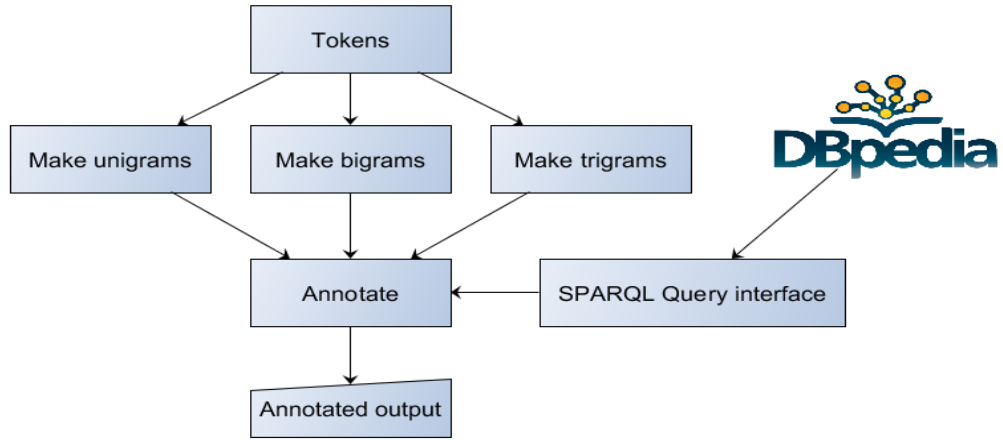

**Figure 2.** Overview of DBPedia workflow

## 2.2. Annotating using WordNet

While doing annotation using WordNet, Marvin is performing also part-of-speech tagging over the inputted text. This is done using OpenNLP part-of-speech tagger based on maximum entropy model for English downloaded from OpenNLP website. Part-of-speech tagging and tokenization is done in that way that for each token, there is also a part-of-speech tag. Using tokens and part-of-speech tags WordNet database is queried. The query returns all the possible senses of the word with a given part-of-speech.

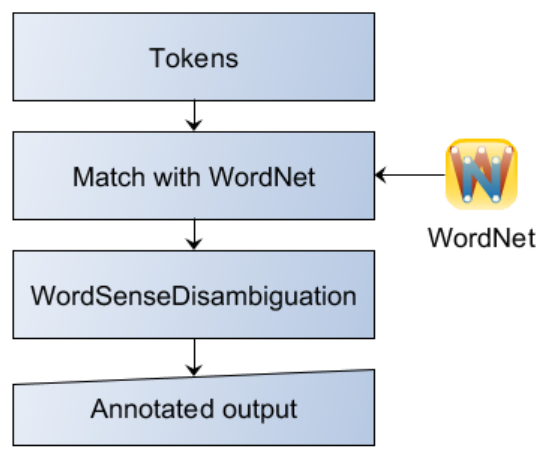

**Figure 3.** Overview of WordNet workflow

Results from the query of WordNet contain senses that are not what text is about. Only one sense of the word is the actual sense in that context. With too many annotations for the senses, the annotations could not be as useful as if they provided the right sense. In order to retrieve only the right sense or a small number of most probable senses we applied word sense disambiguation.

In order to perform word sense disambiguation we modified basic version of Lesk algorithm (Lesk, 1986). The basic idea of Lesk algorithm is to count the number of words in the suroundings of the analysed word and the words that appear in dictionary definition of that term. The idea is very simple and there was, over the years, attempts to improve the algorithm. Issue we found with the algorithm is that for different words, the size of deffinition can be different. Also, the size of the window in which the context is looked for could be different. The ranking should not be the same if the number of matching terms are same for two deffinitions, but one deffinition has more words than the other. Cases like this have to be weightened properly. In order to calculate weights for choosing the right definition, we took 15 words left and right of the current word in the text, if they exist, as a context. The algorithm is calculating for each definition how many words from the definition are appearing in the context of the annotated word. The sum of words appearing in both the context and the definition is divided by the number of the words in definition. By doing this, we are producing a measure which calculates the proportion of the context words in definition. The definition with the largest result is chosen as the meaning of the word. If multiple definitions have same result of our expression, they are all presented as possible definitions of the word.

*2.3. Annotating using UMLS and MetaMap*

Marvin is capable of annotating text using UMLS (Bodenreider, 2004) with the aid of MetaMap (Aronson, 2001). Marvin can send requests for annotations to MetaMap server in case it has the location of MetaMap server. User can configure the location of the server and its port in settings.cfg file. Annotations with UMLS concepts are completely handled by MetaMap and Marvin only enriches these annotations with prevalence information and indexes of the word.

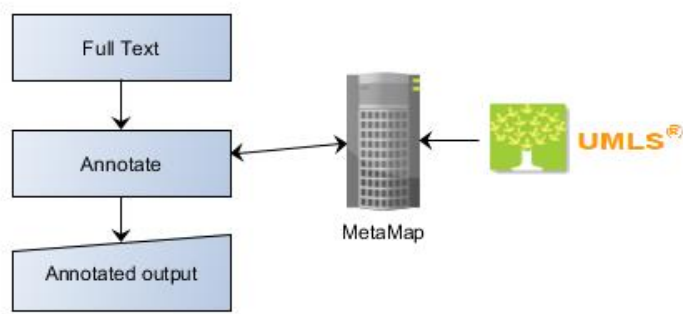

**Figure 4.** Overview of MetaMap workflow

*2.4. Annotating using Simple Knowledge Organization System (SKOS)*

Previously we described annotation with WordNet, DBPedia and MetaMap. These methods are using certain well established vocabularies and they cannot be changed (apart from vendor's updates). However, when performing tasks such as information extraction, sometimes it is necessary to use custom made dictionaries. We have provided a method for users to supply a number of custom vocabularies, which our system will load and annotate text using them. For the vocabulary input format we decided to use Simple Knowledge Organization System (SKOS) format.

Simple Knowledge Organization System is an RDF vocabulary for expressing the basic structure and content of concept schemes, such as thesauri, classification schemes, taxonomies, terminologies, glossaries and other types of controlled vocabularies (Miles, et al., 2005). It is designed and recommended by World Wide Web Consortium as a standard for representing controlled vocabularies (Miles & Bechhofer, 2009). As a W3C standard for representing vocabularies in RDF format, we expect that the format is well developed and adopted in the community.

For the reading of SKOS vocabulary files we used SKOS API that has been designed to work with SKOS models at a high level of abstraction (Jupp, et al., 2009). We have tested the reading of SKOS files created as export from ThManager 2.0, an open source tool for creating and visualizing SKOS (Lacasta, et al., 2013). The concepts from SKOS file are read into two hash maps. The first hash map maps URIs of the concepts into the object that contains other information about that conept, such as broader concept URIs, prefered term and alternative terms. We assumed that each concept contains only one prefered term and can contain zero or more alternative terms. The second hash map maps the words that prefered or alternative labels in SKOS concept contain to the object that fully describes that SKOS concept. It is also important to note that this hash map can contain multiple objects for one term (for example if we have cancer terminology word cancer would contain concepts describing top level cancer concept, as well as more specific concepts such as lung cancer, breast cancer, skin cancer, etc.). We used Google Guava library (Kluever, 2016), which contains multimap objects, which are hash maps with multiple values for one key. The text which needs to be annotated is first transformed to lowercase and broken into the words using tokenizer. For each word, Marvin searchs the hash map that maps words into concepts. If the word is found, our method is taking from the concepts that contain that word prefered and alternative labels and trying to find them in the text. If found, it annotates that part of the text with the associated concept. If the concept contains some broader concept, Marvin will look up for that concept using the second hash map which maps URIs to concepts. If the broader concept is found, that part of the string is annotated also with broader concept. Annotation with brader concepts is continued until the top level is reached. Since annotations are kept seperatly, it is possible to annotate the same word in Marvin with multiple annotations.

*2.5. Provenance*

Provenance comes from French word "provenir" meaning "to come from" and it describes lineage or history of an entity. Provenance metadata are used to provide necessary information to verify the quality of data, validate the data and associate trust value with them (Sahoo & Sheth, 2009). In order to determine what metadata to include into the annotations we consulted The PROV-O ontology (Lebo, et al., 2013). However, PROV ontology is generic and does not contain some specialized descriptions that would be able to unambiguously determine the source of the annotation. We have modified some descriptions, but our provenance is based on PROV-O ontology. We preserve the following metadata:

- Agent name – the name of the agent that annotated data. User needs to define this variable in settings file. Basically it used to describe software or SKOS terminology used for annotation.
- Agent version – the version of the software or terminology used
- Annotation system – the system used for annotation. It is assigned automatically. It can have one of the following values: SKOS, MetaMap, WordNet, and DBPedia.
- Source – Describes the program used for annotation. If Marvin is used it will always return string "MarvinAnnotator"
- Environment description – description of the environment where annotation took place. It can be description of machines, operating systems, software and systems used during the annotation process.
- Date time – The date and time when the annotation took place.
- Location – Geographical location where the data was annotated

Agent name, version, environment description and location are described by the used in Marvin's settings file. The rest of the provenance data is generated automatically.

## 3. Summary and Future Works

In this report we presented Marvin, an application that is able to annotate text using different semantic data resources, including MetaMap, WordNet, DBPedia and custom thesauri represented in SKOS format. Marvin can be used as command line tool or as a Java library. We assume that making it a library can benefit many users who want to use some of its features, however, have control over what they are using and allowing them to customize the output for their purposes.

The motivation for developing Marvin came from the need of a flexible text annotator for annotating and normalizing tables. We needed annotated and normalized text in tables for information extraction purposes (Milosevic, et al., 2016). However, the tool is built to be universal for any kind of text.

For the future we may extend our software to support more formats and annotation systems. Also, we have an idea to extend support of SKOS, so SKOS thesauri can contain regular exptressions (regex) as the alternative terms for some concepts. This could be helpful when producing thesauri, since one do not need to include all flections of the word. Also, value presentation hierachies could be represented this way. These alternative terms would have some flag to denote that they use regular expression (for example "[!re]" on the start, or if agreed with W3C recommendation board a new attribute can be introduced ). However, challenge for this approach can be the speed of processing. Since regular expressions cannot be used anymore as hash maps, the algorithm would need to iterate over all stated regular expressions and try to find them in text. This sequential iteration will cause a significant reduction of algorithm's speed. However, we may introduce a list that would separately process regular expressions and remain with the current approach for non regular expression statements.

**4. Accessibility**

Marvin is an open source project, which source code and binaries are freely available on GitHub on the following location: https://github.com/nikolamilosevic86/Marvin.

The website with additional information on how to set up and run Marvin annotator is available on the following address: http://nikolamilosevic86.github.io/Marvin/

# Technical report